\pdfoutput=1

\documentclass[11pt]{article}

\usepackage[preprint]{acl}

\usepackage{times}
\usepackage{latexsym}
\usepackage{url}
\usepackage{multirow, hhline, graphicx, subcaption, helvet}
\usepackage[most]{tcolorbox}
\usepackage{colortbl}
\usepackage{float} 
\usepackage[T1]{fontenc}

\usepackage[utf8]{inputenc}

\usepackage{microtype}

\usepackage{inconsolata}

\usepackage{booktabs}

\usepackage{xspace}

\usepackage{csquotes}

\usepackage{rotating}

\usepackage{array}
\usepackage{amsmath}

\newcommand{\instruction}{\textsc{instruction}\xspace}
\newcommand{\inp}{\textsc{input}\xspace}
\newcommand{\out}{\textsc{output}}
\newcommand{\expl}{\textsc{explanation}}
\newcommand{\llinstruct}{\textsc{ll-instruct}\xspace}
\newcommand{\selfinstruct}{\textsc{self-instruct}\xspace}
\newcommand{\instruct}{\textsc{instruct}\xspace}

\makeatletter
\newcommand{\printfnsymbol}[1]{%
  \textsuperscript{\@fnsymbol{#1}}%
}

\title{\llinstruct: An Instruction-tuned model for English Language Proficiency Assessments}

\author{Debanjan Ghosh\thanks{~~Equal Contribution.} \\
Educational Testing Service \\
  \texttt{dghosh@ets.org} \\\And
  Sophia Chan\printfnsymbol{1} \\
  Educational Testing Service, Canada  \\
  \texttt{schan@etscanada.ca} \\}

\begin{document}
\maketitle
\begin{abstract}

We present \llinstruct: An 8B instruction-tuned model that is designed to generate content for English Language Proficiency Assessments (ELPA) and related applications. Our work involves creating a new dataset of 70K instructions and explanations in the ELPA domain and using these to fine-tune Llama-3 8B models (SFT) of different sizes (e.g., SFT-17K, SFT-50K and SFT-70K). Human evaluations are conducted over unseen instructions to compare these SFT models against SOTA models (e.g., Dolly-2, Mistral, Llama-3 base version, and GPT-3.5). The findings show although all three SFT models perform comparably, the model trained on largest instruction dataset  --  SFT-70K - leads to the most valid outputs ready for assessments. However, although the SFT models perform better than larger model, e.g., GPT 3.5 on the aspect of explanations of outputs, many outputs still need human interventions to make them actual ready for real world assessments.

\end{abstract}

\section{Introduction} \label{section:intro}

Instruction tuning---or multitask prompted finetuning---is an area within Natural Language Processing (NLP) that has led to state-of-the-art performance across a variety of tasks in recent years \cite{ouyang2022training,sanh2022multitask,wang2022self,peng2023instruction}. Broadly, this involves training a pre-trained Language Model (LM) using 
<\instruction, \out> pairs where the \instruction describes the task in natural language and \out\xspace  is the desired outcome. As mentioned in \newcite{peng2023instruction}, current research primarily focuses on two sub-areas: scaling instruction-tuning models using more training data and resources and/or utilizing large amounts of human-annotated data for training. 
However, it is known that (a)  gathering human-annotated data is expensive and time-consuming and (b) instruction-tuned models that are specialized for specific NLP tasks tend to struggle when applied to a diverse range of tasks \cite{zhang2023instruction}. 

\begin{table*}
\small
\centering
\begin{tabular}{p{0.16\linewidth}p{0.77\linewidth}}\toprule
Instruction & \hangindent=0.5em Reply to the following email and express you can't attend the meeting due to a personal issue. \\
 Input & \hangindent=0.5em We have scheduled a meeting at 4 PM tomorrow to discuss our project. Please confirm your availability.  \\
Output & \hangindent=0.5em Thank you for the information. Unfortunately, I won't be able to attend the meeting tomorrow due to a personal issue. However, I'll make sure to catch up on the meeting notes. \\
Explanation & \hangindent=1.5em The response is polite and appropriate as it expresses the inability to attend the meeting due to a personal issue and includes an offer to catch up on what was discussed. \\

\midrule
Instruction & \hangindent=0.5em Read the email. Create a question that asks about the scheduled date for the product launch.\\
 Input & \hangindent=0.5em Email: Dear John, Just a quick reminder about the product launch happening next week. The team is keenly getting ready for it, and we appreciate all the hard work and dedication you've shown to make this successful. Best, Jane\ \\
Output & \hangindent=0.5em When is the product launch scheduled? \\
Explanation & \hangindent=1.5em The answer correctly specifies a question about the product launch date mentioned in the email. \\
\midrule
 \end{tabular}
 \caption{\label{table:ll_data} A few examples of instructions that were generated during the iterative prompting stage. This stage is described in Section \ref{subsection:autodata}.}

\end{table*}


In response to these shortcomings, several recent models have been trained using a  diverse set instructions generated via a semi-automated method. \newcite{wang2022self}  collected a small set of manually-written <\instruction, \inp, \out> examples and then used the set of examples to prompt GPT-3 \cite{brown2020language} to generate a larger set of more diverse instructions. The authors then fine-tuned GPT-3 using the generated tuples. This approach is named as \selfinstruct as the final model is trained on \emph{self-generated} instructions. 




Inspired by the aforementioned approach, in this paper we introduce \textbf{(L)anguage (L)earning \instruct (henceforth, \llinstruct)}: an instruction tuned model specifically designed for English Language Proficiency Assessments (ELPA) and other related applications in the Educational Technology (EdTech) domain. The Language Learning market is experiencing a significant growth with a projection of surpassing 100 billion USD in the coming years.\footnote{\url{https://www.meticulousresearch.com/pressrelease/792/language-learning-market-2030}} This coincides with the rise in the utilization of large pretrained LMs, with multiple organizations adopting automated content generation for ELPA such as Duolingo   \cite{settles2020machine, burstein2021theoretical} and Cambridge Assessment English \cite{galaczi2023english}.\footnote{We use LLM and pretrained LMs interchangeably.} 



There are several distinguishing aspects of \llinstruct that set it apart from other instruction-tuned models such as Alpaca \cite{alpaca}, WizardLM \cite{xu2024wizardlm}, LLaMA-GPT4 \cite{peng2023instruction},  RoleLLM \cite{wang2024rolellm} and so on. \textbf{First}, the authors manually write 130 seed instructions in the form of <\instruction, \inp, \out> tuples based on publicly available test items from standard ELPA. These are tailored to assess skills such as reading, speaking, listening, and writing, which are crucial for language learning and testing (Section \ref{subsection:seeddata}).\footnote{This four skills approach is widely adopted by ELPA and the language teaching community. While \citet{powers2010case} advocates  assessing these skills individually, \citet{hinkel2010integrating} notes it is possible to integrate them in pedagogy.} \textbf{Second}, we use the seed tuples to iteratively prompt GPT-4 \cite{gpt4technicalreport} to generate more data, i.e., 70K <\instruction, \inp, \out> tuples. \textbf{Third}, we use another LLM as a discriminator to reject any generated instructions that do not contribute to ELPA, such as, ``What is the capital of Australia?'' in a separate evaluation stage, similar to how label-correctness in computed in \citet{perez-etal-2023-discovering}. \textbf{Fourth}, an  \expl\xspace  for each output is generated to assist in understanding the reasoning behind the outputs making our outputs as  <\instruction, \inp, \out, \expl>. This feature can be highly valuable for test designers and individuals taking practice tests to understand the rationale behind the outputs. Table \ref{table:ll_data} contains two instruction tuples that are suitable for ELPA and generated during the iterative prompting stage (Section \ref{subsection:autodata}).  \textbf{Fifth}, we fine-tune Llama-3 8B \cite{meta2024introducing} models with different dataset sizes: 17K, 50K, and 70K. 
\textbf{Finally}, we conduct a comprehensive human evaluation for 200 unseen instructions. Several pre-trained models are evaluated alongside the fine-tuned models: Dolly-2 8B \cite{DatabricksBlog2023DollyV2}, Mistral 7B \cite{jiang2023mistral},  Llama-3 8B \cite{meta2024introducing}, and GPT-3.5. The fine-tuning process is described in Section \ref{section:exp}.

The results are detailed in Section \ref{section:performance-eval}. We find that although all three SFT models are comparable,   SFT-70K produced most number of outputs that are valid and ready for use in ELPA. Specifically, both SFT-70K and GPT-3.5 produce above 60\% valid and ready outputs  exceeding Dolly-2, Mistral, and Llama-3 base version. We also notice, SFT-70K also generated the highest number of usable explanations while this number was much lower for non-fine-tuned models (SFT-70K: 80.5\%, GPT-3.5: 42\%). However, in many cases, subject matter expert (SME) revisions are still necessary to adapt the outputs for assessment readiness (see Section \ref{subsection:modelcompare}).\footnote{Items in standardized ELPAs include specific constructs related to grammar, semantics, and pragmatics, typically authored by assessment developers.}  All datasets and models will be released upon acceptance.

\begin{figure*}[t]
\begin{center}

\includegraphics[width=.9\textwidth]{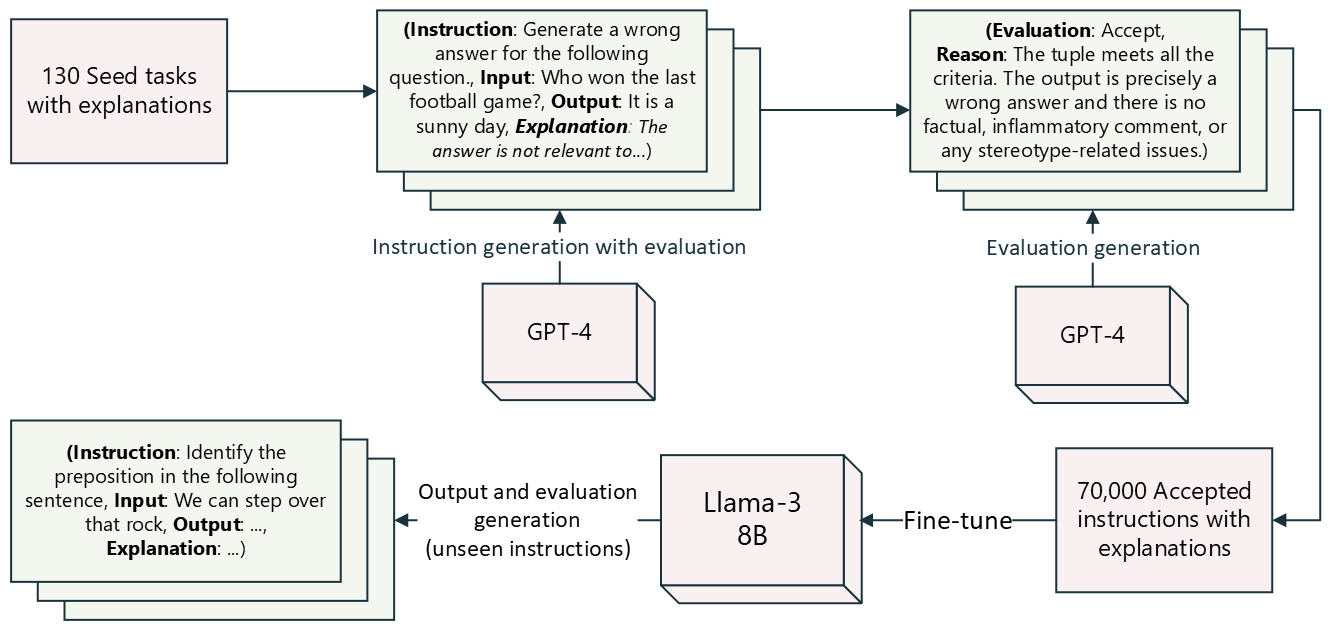}

\end{center}
\caption{\label{fig:sample-figure}Overview of our proposed Instruction data generation and filtration strategies} 
\end{figure*}\label{section:data}

\section{Datasets} 

\begin{table*}[!t]
\small
\centering
\setlength{\tabcolsep}{3pt}
\begin{subtable}{1\linewidth}
\begin{tabular}{p{0.5\linewidth}p{0.5\linewidth}}\toprule
\textbf{Original Item} & \textbf{Seed Instruction} \\
\hline
Choose the best response. 
\newline \newline Statement: who is your favorite tennis player? 
\newline \newline Answers: [`play the song again', `my favorite color is green', `that sounds like fun', \textbf{`I don't like any sports'}] & 
\instruction: Write an indirect response to the following question.
\newline \inp: who is your favorite tennis player?
\newline \out: I don't like any sports. 
\newline \expl: Reply indirectly addresses the question by stating a general disinterest in sports rather than specifically identifying a favorite tennis player. \\

\cline{2-2}
& \instruction: Write three wrong answers to the question.
\newline \inp: who is your favorite tennis player?
\newline \out: [`play the song again', `my favorite color is green', `that sounds like fun'] 
\newline \expl: The given answers are either unrelated (e.g.  ``play the song again''), changing the subject (e.g. talking about a favorite color), or expressing a general sentiment (e.g. commenting on an activity). \\



\hline

 \end{tabular}
 \end{subtable}

\caption{\label{table:seed} Original ELPA item (first column) and extracted seed instructions (second column) from the items. The \textbf{bold} selections present the correct choices. Original items are lightly edited for anonymity.}
\end{table*}

In this section we describe in detail our approach of collecting training data to build the \llinstruct model. This includes: curating seed instructions, generating diverse instructions with a LLM, and filtering the generated instructions. The full flow is shown in Figure \ref{fig:sample-figure}.

\subsection{Curating Seed Instructions} \label{subsection:seeddata}

To start, we curate a set of seed instructions so that they can be the basis for generating additional instructions. Note, a key differentiating factor in our work is that our instructions are solely based on ELPA as applicable in the field of EdTech.

The authors, who are familiar with standard ELPA, begin by converting publicly available ELPA items into instruction examples.\footnote{In the assessment domain, an ``item'' is a term commonly used to denote a stimulus accompanied by a question/answer set. Many EdTech companies conducting ELPAs globally provide sample tests online.} In most of the cases, we split a single item into multiple instructions because often many ELPA items contain multiple sub-items. A common format involves a stimulus (e.g., source text) followed by multiple-choice questions (MCQs) that include a question, correct option, and incorrect options. An instruction can be created for each part or encompass multiple parts.
 ELPA test designers may be a significant user group for the \llinstruct model, and we foresee them employing multiple <\instruction, \inp, \out> tuples to construct a single ELPA item. 

 

To illustrate the process better let us explore the example from Table \ref{table:seed}. The first column, ``Original Item'', is an item taken directly from a sample standard ELPA (available online). Here, the test taker is required to choose the most appropriate response to reply the statement. In the second column,``Seed Instruction'', we created two <\instruction, \inp, \out> tuples from the item. Given the best response``I don't like any sports'' is an \emph{indirect} answer to the question asked we create a seed instruction to (e.g. , ``write an \emph{indirect} response \dots'') to reflect that. Likewise, we create the second instruction that focuses on generating ``three wrong'' answers, commonly known as the distractors that usually designed to divert the test takers from selecting the correct answer.

Each tuple also includes an \expl\xspace, which proves helpful in understanding the reasoning behind each output. Seed explanations ( Table \ref{table:seed}) are written by the authors of the paper. Also note, it is not always necessary to generate more than one seed instruction from a test item. The goal is to create a variety of seed instructions that capture all the different components we wish to generate using the \llinstruct model in the end.




In total we compiled 130 seed instructions inspired by the standard ELPA. To increase variation, we occasionally  place the input text within the \instruction  and at other times within the \inp section. We also adopt linguistic variations in the instructions by using different phrases for the same type of tasks, like ``generate the answers,'' ``create a sentence,'' ``write the answer,'' and so forth to mitigate the decline of linguistic diversity found in LMs trained on synthetic text  \cite{guo-etal-2024-curious}. 


\subsection{Automated ELPA Instruction Generation} \label{subsection:autodata}

Similar to \citet{wang2022self} we generate new instructions via a bootstrapping method. In each step, we include four <\instruction, \inp, \out, \expl> instruction tuples in the prompt: three seed tuples and one model-generated tuple. To promote diversity, we split the 130 seed tuples into two categories: \textbf{(a) short tasks} (e.g., grammar correction, convert a passive sentence into active, etc.) and \textbf{(b) long tasks} (e.g., write an email, write a short conversation, etc.) and then randomly choose either two short tasks and one long task or vice-versa in the prompt. Then, GPT-4 is prompted to generate ten new tuples, corresponding to ten new instructions. 

Refer to Section \ref{sec:prompt-template-1} in the Appendix section for the prompt template. We have included a few requirements in the prompts that are intended to guide the model. For example, we ask that the new instructions be relevant to ELPA and not involve generating code or solving arithmetic problems.


\subsection{ELPA Instruction Data Filtration} \label{subsection:dataeval}
Despite explicitly prompting against it, we notice that the model sometimes generate tasks about factual information, such as ``What is the capital of Australia?''. As mentioned before, this kind of task data is not beneficial for ELPA and we aim to exclude such data in our fine-tuning. Evaluation of LLM outputs guided by another LLM has been shown to be effective \citep{chiang-lee-2023-large}, thus, to remove factual data from the \textbf{Automatic ELPA Instruction Data Generation} round, we use another GPT-4 model.


We write a new prompt (refer to Section \ref{sec:prompt-template-2} in the Appendix for the complete template) that includes examples of both factual tasks  that we intend to exclude and non-factual tasks that we want. After conducting a small pilot test, we found that approximately 7\% of non-ELPA tasks were produced by the bootstrapping step (Section \ref{subsection:autodata}), which have now been flagged and removed. Furthermore, we enhance ELPA instruction data quality through standard filtering and postprocessing, removing instructions with irrelevant terms, such as, video, image, graph, flowchart, etc. To maintain diversity, we avoid adding instructions too similar to existing ones, using a ROUGE-L metric to ensure no two instructions exceed a 0.75 similarity score (this was set empirically after tuning).

Following the completion of all filtering processes, we are left with 70K instructional data. In the next section we present an evaluation of the generated data and describe the contents.


\section{Evaluation of \llinstruct  Data} \label{section:overview}




Our work focuses on the quality of automatically generated \llinstruct data, crucial for tailoring SFT models to ELPA. Unlike \newcite{wang2022self} and related studies, we conduct a large-scale evaluation to ensure the instructions' relevance and suitability for English language assessments.

We randomly selected 250 generated instruction tuples and carried out the evaluation in two stages. First, we classify the instruction tuples by language \textbf{category} (e.g., grammar, semantic, etc.) and language \textbf{skills} (e.g., speaking, writing) to demonstrate the types of instructions included in the dataset (Section \ref{subsection:langcat}). Next, we further focus into specific aspects of the instruction tuples, such as, output correctness, quality of explanation, etc. (Section \ref{subsection:instqual}). 

\subsection{First Evaluation Task: Language Category and Skills} \label{subsection:langcat}
\begin{figure*}
\centering
\begin{subfigure}{.45\textwidth}
  \centering
  \includegraphics[width=\linewidth]{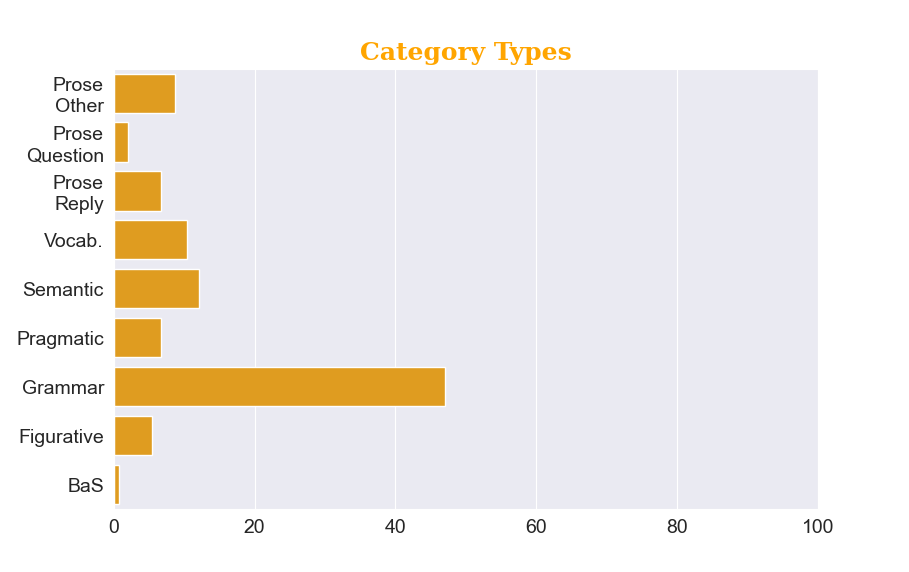}
  \caption{\label{fig:categories}}
\end{subfigure}%
\begin{subfigure}{.42\textwidth}
  \centering
  \includegraphics[width=\linewidth]{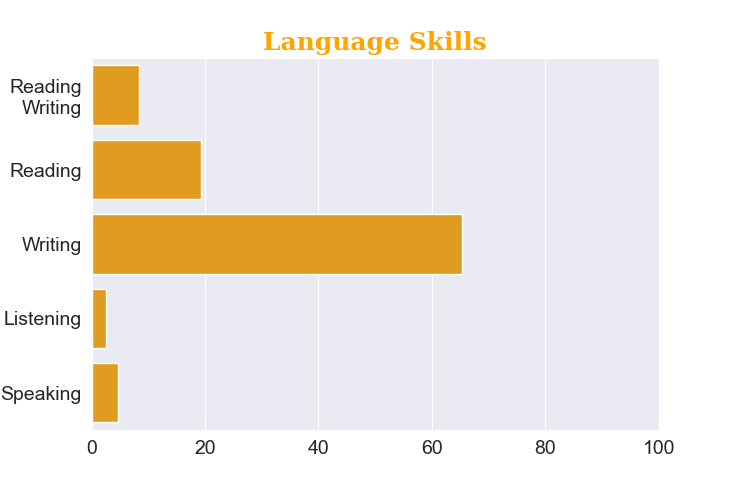}
  \caption{\label{fig:tasks}}
\end{subfigure}
\caption{Percentage of \llinstruct data by language categories(a) and skills(b).}
\label{fig:test}
\end{figure*}

The language \textbf{category} of an instruction specifies the type of linguistic knowledge that the <\instruction, \inp, \out> and the resulting ELPA item probes for. For example, some categories are grammar, vocabulary, semantic, pragmatic, and prose (i.e. prose writing). Likewise, we also categorized the instruction tuples to language \textbf{skill} such as reading and writing. Authors of this papers first conducted a pilot annotation task to determine the main categories and then jointly annotated the remaining examples. Figure \ref{fig:categories} and Figure \ref{fig:tasks} present the main categories and  skills  identified from the example set, respectively.

In terms of the observed categories, instructions related to grammar (e.g., ``rewrite the question in reported speech'', ``identify the preposition in the sentence'' etc.) are frequently encountered. Additionally, there are numerous instructions pertaining to prose writing that we have further categorized into ``Prose Question'' (e.g., ``write question(s) based on the passage''), ``Prose Reply'' (e.g., ``write a reply to the dialog''), and ``Prose Other'' that contains many different writing tasks such as  ``write a opinionated argument on topic \emph{x}''. We notice many  pragmatic (e.g., ``identify the informal words''), and figurative instructions (e.g., ``write a simile about a everyday item'', ``identify the implied metaphor in the statement'') too. Besides these, interestingly, instructions aligned with language learning assessment such as Build a Sentence (abbreviated as BaS in Figure \ref{fig:categories}), such as ``compose a sentence with the following words'' also appear as well. 

Regarding the types of language skills, we observe that the majority can be classified as writing tasks. This includes most grammar and semantic instruction categories, in addition to the clearly defined prose category. We also identified some instructions as both reading/writing tasks  (e.g. ``read the following sentence and paraphrase it in the past simple tense'', ``read the statement and suggest an alternative word ...''), which can be seen as two combined tasks. 

\subsection{Second Evaluation Task: Instruction Quality}
\label{subsection:instqual}

Here, we focus onto the following aspects of the instructions. 
\begin{itemize}
    \item \textbf{Validity}: whether the example is valid and ready to appear in an English language assessment. We provide three options: \emph{valid and ready} for assessment, only \emph{valid} (i.e., needs some editing), and \emph{invalid}.
    \item \textbf{Instruction type}: whether the instruction is \emph{factual} or \emph{not factual}.
    \item \textbf{Input faithfulness}: does the input \emph{matches} or \emph{not matches} to the instruction. 
    \item \textbf{Output correctness}: whether the output is \emph{correct} (based on the instruction) or not.
    \item \textbf{Quality of explanation}: does the explanation justify the output? We provide four options: \emph{yes}, \emph{weak yes}, \emph{weak no}, and \emph{no}.
\end{itemize}

We recruit ten expert annotators, each with a background in linguistics, computer science, and EdTech experience, to evaluate 50 instruction tuples per pair of annotators. We measure Krippendorf's $\alpha$ \cite{krippendorff2011computing} on each aspect for each pair of annotation and then report the average $\alpha$. They are: 0.49 for \textbf{Validity} (moderate agreement), 0.93 for \textbf{Instruction type} (almost perfect agreement), 0.67 for \textbf{Input faithfulness} (substantial agreement),  0.78 for \textbf{Output correctness} (substantial agreement), and 0.52 for \textbf{Quality of explanation} (moderate agreement). We focus into \textbf{Validity} and \textbf{Quality of explanation} aspects where the agreement is comparative lower than the other aspects. Regarding Validity, most of the disagreements occur between choosing the instructions as \emph{valid and ready} vs. \emph{valid}. We notice, in case of instructions like ``identify the \emph{tense/voice/verb type \dots} of the sentence,'' some annotators indicated that providing a list of possible options is typical in an actual assessment rather than directly asking for identification. Similarly, for instructions like ``provide a \emph{synonym/alternate ending/\dots},'' some annotators critiqued them as being too open-ended, suggesting that these usually avoid such ambiguity or provide options to choose from. Likewise for \emph{Quality of explanation} aspect annotators sometimes disagree on whether an explanation is sufficient (``yes'') or need human editing (``weak yes'').




\section{Experimental Details} \label{section:exp}
The experiment involves supervised fine-tuning (SFT) a Llama-3 8B model.
The design choice to use a small 8B model is driven by two primary motivations: (a) to evaluate how effective a small SFT model can be for language learning applications, and (b) to ensure fast inference and moderate GPU requirements, thereby lowering the barrier to trying these model(s).

\paragraph{Fine-tuning Llama-3 8B}
Llama-3 8B was fine-tuned on subsets of the \llinstruct data of size 17K, 50K, and 70K.\footnote{We chose the 17K and 50K partitions randomly from the total of 70K instruction tuples that are generated.} Each <\instruction, \inp, \out, \expl> tuple was joined into one example using the following template: 
\begin{displayquote}
\small
Below is an instruction that describes a task. Write a response that appropriately completes the request. \#\#\# Instruction: \instruction \#\#\# Input: \inp \#\#\# Output: \out \#\#\# Explanation: \expl
\end{displayquote}

Huggingface repository is used to perform the SFT.\footnote{\url{https://github.com/huggingface/trl}}
Parameter specifications can be found in the full training command in Section \ref{sec:train-command} in the Appendix.

\paragraph{Inference on Test Dataset} 


To evaluate the quality of the SFT models (SFT-17K, SFT-50K, and SFT-70K) they are  compared against several SOTA models (only pretrained without any additional fine tuning): base Llama-3 8B \cite{meta2024introducing}, GPT-3.5 \cite{brown2020language}, Mistral 7B \cite{jiang2023mistral}, and Dolly-2 8B \cite{DatabricksBlog2023DollyV2}.
The prompt for inference is similar to the one used for fine tuning, except we start generating at the output:

\begin{displayquote}
\small
Write the output by following the instruction and the input, and then include an explanation for why the output is appropriate given instruction and input. Include a separator token `\#\#\#` before the explanation.\\
\#\#\# Instruction: \instruction \#\#\# Input: \inp \#\#\# Output: 
\end{displayquote}


We selected an unseen batch of 200 instructions to do the comparison, where the instructions are ranged over diverse tasks such as grammar, figurative language and prose.  

\section{Human Evaluation of Model Performance}\label{section:performance-eval}
For each test instruction, the authors jointly evaluated the output from each of seven models, i.e., a total of 200x7, 1400 outputs. We re-use the rubric that was used to evaluate the quality of \llinstruct in Section \ref{subsection:instqual}. We assessed dimensions of \textbf{Validity}, \textbf{Output correctness}, and \textbf{Quality of explanation}, but omitted \textbf{Instruction type} and \textbf{Input faithfulness}  dimensions due to the non-factual nature of almost all instructions and and the lack of dedicated input entries in the test set, respectively.

\begin{figure}[h]
   \centering
   \includegraphics[width=1.0\linewidth]{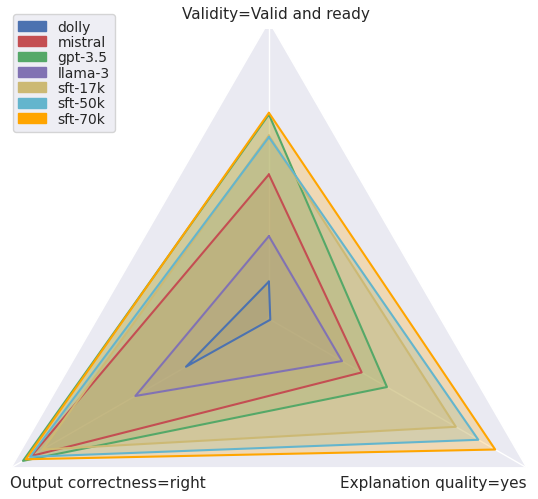}
   \caption{Comparison of human evaluation results across all seven models on three dimensions (Validity, Output correctness, and Explanation quality).}
   \label{fig:result_spider}
\end{figure}


\subsection{Comparison Between All Models} \label{subsection:modelcompare}

An overview of how the models fared on each dimension can be seen in Figure \ref{fig:result_spider}. Table \ref{tab:result-raw-numbers} summarizes the results, showing the percentage of data in each category. Additional tables that show the win-rate and tie-rate between each model can be found in Section \ref{subsection:win-tie-rate} in the Appendix. Let us examine the results for each of these three dimensions in turn. 

\newcolumntype{P}[1]{>{\centering\arraybackslash}p{#1}}

\begin{table*}[!t]
    \small
    \centering
    \begin{tabular}{l|P{1.cm}|P{1.cm}|P{1.cm}|P{1.cm}|P{1.cm}|P{1.cm}|P{1.cm}|}
         & \rotatebox{60}{Dolly-2} & \rotatebox{60}{Mistral} & \rotatebox{60}{GPT-3.5} & \rotatebox{60}{Llama-3} & \rotatebox{60}{SFT-17K} & \rotatebox{60}{SFT-50K} & \rotatebox{60}{SFT-70K} \\
         \hline 
        \textbf{Validity} &  &  &  &  &  &  & \\
        \arrayrulecolor{gray}\hline 
        \hspace{3mm}\textit{valid and ready} & 11.50 & 44.50 & 63.00 & 25.50 & 56.50 & 56.00 & 63.50 \\
        \hspace{3mm}\textit{valid} & 16.50 & 41.50 & 23.50 & 22.50 & 24.00 & 29.00 & 22.50 \\
        \hspace{3mm}\textit{invalid} & 72.00 & 14.00 & 13.50 & 52.00 & 19.50 & 15.00 & 14.00 \\
        \hline
        \textbf{Output correctness} &  &  &  &  &  &  & \\
        \hline 
        \hspace{3mm}\textit{right} & 29.50 & 84.50 & 87.50 & 47.50 & 80.50 & 85.00 & 86.50 \\
        \hspace{3mm}\textit{wrong} & 70.50 & 15.50 &  12.50 & 52.50 & 19.50 & 15.00 & 13.50 \\
        \hline
        \textbf{Explanation quality} &  &  &  &  &  &  & \\
        \hline
        \hspace{3mm}\textit{yes} & 0.50 & 33.00 & 42.00 & 26.00 & 66.50 & 74.50 & 80.50 \\
        \hspace{3mm}\textit{weak yes} & 1.00 & 32.50 & 9.00 & 17.00 & 9.50 & 7.00 & 5.50 \\
        \hspace{3mm}\textit{weak no} & 3.00 & 5.00& 1.00 & 3.50 & 1.00 & 0.50&0.00 \\
        \hspace{3mm}\textit{no} & 95.50 & 29.50 & 48.00 & 53.50 & 23.00 &  18.00 & 14.00 \\
        \hline
    \end{tabular}
    \caption{Results from human evaluation of model performance. Proportion of a models' output along three dimensions: Validity, Output correctness, and Explanation quality.}
    \label{tab:result-raw-numbers}
\end{table*}

\paragraph{Validity} SFT-70K and GPT-3.5 have the highest number of \textit{valid and ready} generations where SFT-70K is marginally better (63.5\% vs. 63\%, see Table \ref{tab:result-raw-numbers}) followed by SFT-50K and SFT-17K almost equally. In contrast, Mistral, Llama-3, and Dolly-2 have fewer than 50\% valid and ready generations, with Mistral leading at 45\%, followed by Llama-3 at 26\%, and Dolly-2 at 12\%.  


\paragraph{Output Correctness} GPT-3.5 has the highest number of \textit{right} generations (87.5\%), followed closely by SFT-70K (86.5\%), whereas the two remaining SFT models, SFT-50K and SFT-17K perform similar to Mistral (all between 80-85\%).  The remaining two models, Llama-3 (base model) and Dolly-2 have fewer than 50\% right generations. 

\paragraph{Explanation Quality} The explanations are rated on a four-category scale, with \textit{yes} being the best and \textit{no} being the worst. SFT-70K has the most explanations in the \textit{yes} category (80.5\%), followed by SFT-50K (74.5\%), then SFT-17K (66.5\%). Perhaps the SFT models showcase their usefulness the most for this aspect, given, the closest best explanations are from GPT-3.5 (42\%), which is 38.5\% lower than SFT-70K model's performance.

\subsection{Qualitative Error Analysis}

We conducted a thorough error analysis of all the model outputs and highlighted specific characteristics here. 

\paragraph{Verbose outputs and explanations}  Most often for GPT-3.5, Mistral, and Llama-3, while the output and explanation match the specification, an excessive amount of words and description is used. Consider the following instruction:

\begin{displayquote}
    \small
    \instruction: Write a [passage type = email] from [org = Blue Sky Airlines] to [person = Passenger] regarding their lost luggage case. Explain the steps the company is taking to locate the luggage and reassure them. The length of the answer should be around 100 words. \\
    \inp: noinput
\end{displayquote}

GPT-3.5 produced a 212-word email, while SFT models created emails around 100 words. Without specified word limits in instructions, GPT-3.5 often wrote very long responses, sometimes reaching 250-300 words. Llama-3 (base) and Mistral also frequently created longer responses.

Verbose outputs are difficult to assess and edit, so they are not preferred. Our evaluation, however, overlooks verbosity unless it makes the output ineligible, since it does not impact Validity or Output Correctness.

\paragraph{Explanations are often missing} Even if we prompt the models to include explanation for the output, we find that often the explanations are missing. For instance, we notice that explanations are missing from 95\% of Dolly-2 generations, 55\% of Mistral generations, 51\% of GPT-3.5 generations, and 10\% of Llama-3 generations. On the contrary, only  less than 1\% of any fine-tuned model (SFT-17K, SFT-50K, and SFT-70K) is missing the explanation. 



\paragraph{Formatting errors of outputs}

Formatting errors in the generation can hinder the full automation of ELPA item generation. Common formatting issues include a numbered list being returned when only one item is requested, typically by the base Llama-3 model.
Other problems involve the separator token \#\#\# (between the \out\xspace and \expl) being misplaced, repetition of the instruction in the output, and ignoring the word limit specifications, and so on. However, SFT models exhibit less frequently such formatting errors. 


\paragraph{Outputs are often in the proximity but do not follow the instruction exactly} In general, we often notice errors where the models' interpretation of the instruction is close, yet it does not adhere to the request by missing some part of the instruction or simply adding extra information (hallucination). 



Consider the following instruction:
\begin{displayquote}
    \small
    \instruction: Paraphrase the sentence. \\
    \inp: Nobody knew how much time she spent training for the Olympic Games.
\end{displayquote}
Dolly-2 generates an output which contains a close paraphrase, yet, add extra hallucinated information imagining the content is regarding a freestyle skier.  
Likewise, Llama-3 produces an output: \textit{``She trained a lot for the Olympic Games''} which fails to fully convey the input's meaning. Interestingly, the SFT models excel at accurately capturing this precise instruction (refer to Section \ref{proximity-example} in the Appendix for all outputs).



\paragraph{Tasks involving figurative language are difficult} Grammar, vocabulary, and simpler prose tasks tend to be cases where most models produce valid output. On the other hand, figurative language tends to be difficult for most models. Even the SFT-50K model sometime misses generating content that includes a figurative type, e.g. an idiom (See Example \ref{hard-example} in the Appendix).



\section{Related Work} \label{section:related}

Recent studies have demonstrated that  LMs can effectively follow language instructions when fine-tuned using human-annotated datasets that pair instructions with outputs \cite{weller-etal-2020-learning,sanh2022multitask,peng2023instruction}. However, to address the reliance (bottleneck) on large-scale human annotations, researchers such as \cite{ouyang2022training,wang2022self} have developed general-purpose LMs designed to follow diverse sets of instructions. Our research is closely aligned with \cite{wang2022self}, which presented the idea of self-instruct (i.e., an iterative method for creating new instructions and outputs to enhance fine-tuning) and then taken by models such as Alpaca \cite{alpaca}, WizardLM \cite{xu2024wizardlm}, LLaMA-GPT4 \cite{peng2023instruction},  RoleLLM \cite{wang2024rolellm}. What sets our work apart is our specific focus on the language learning and assessment domain, where all instructions pertain to language categories like grammar, vocabulary, semantics, and pragmatics, and are spread across language skills such as reading and writing.

Similar to our approach, \textit{Humpback} \cite{li2023self} utilizes a collection of curated seed instructions to generate new ones. However, a key distinction is that \textit{Humpback} derives its outputs from existing web corpora, whereas we generate all components of the <instruction, input, output> tuple using datasets generated by models like GPT-4. Although our work relates to the self-training literature, which typically defines a specific target as noted by \cite{wang2022self}, our approach is different. Despite our instructions being focused on language assessment, they exhibit wide diversity across various instruction types. Lastly, our research aligns with the concept of distillation \cite{hinton2015}, as we extract new instructions from a teacher model (in this case, GPT-4). Additionally, we employ a separate language model as a discriminator to eliminate factual inaccuracies and non-ELPA instructions.

\section{Conclusion and Future Work} \label{section:conclusion}
We compiled a set of instruction seed data consisting of <\instruction, \inp, \out, \expl> tuples designed for item generation in ELPA. Using these seed instructions, we prompted GPT-4 to generate a much larger dataset of instruction tuples (\llinstruct) for ELPA domain. Subsequently, we fine-tuned Llama-3 model using with different partitions (17K, 50K, and 70K) of the \llinstruct data. We compare the performance of the fine tuned models against various LM baselines including Dolly-2, Mistral, GPT-3.5, and Llama-3 (pretrained).  The fine-tuned versions consistently demonstrated superior performance in terms of output validity, correctness, and explanation quality (Section \ref{subsection:dataeval}).

Our detailed error analysis identified common issues across the models, such as verbose responses (often from GPT-3.5), misunderstanding of instructions (often by Dolly-2), and formatting errors (e.g., Dolly-2 and Mistral). More importantly, we observe that while the fine-tuned Llama-3 models produced approximately 60\% of outputs that were immediately test-ready, about 20-30\% still required manual adjustments by Subject Matter Experts (SMEs) in the language learning field. This suggests that a combined human-AI approach would be most effective for advancing ELPA task designs.

For future work, we plan to improve our SFT model by aligning with human preference, e.g., DPO \cite{rafailov2023direct}, specifically related to the unique language learning domain. We also plan to align the trained models to specific attributes (e.g. quality of explanation, output correctness) by post-hoc merging of parameters (similar to  \newcite{jang2023personalized}).

\section{Ethics}
The risks and harms of language models are well-documented. \citet{bender2021dangers} provides an overview, including: environmental and financial cost; unfathomable training data leading to encoded biases that reflect the dominant/hegemonic view; coherent output being mistaken as true knowledge. 

This work uses GPT-4 (1.76 trillion parameters) for dataset generation. In addition to increased water consumption and carbon emissions \citep{strubell2020energy, george2023environmental} when using a larger model, there is the risk of including harmful biases and misinformation in both the training data and in the fine-tuned models. The data was spot-checked and filtered using another LLM to remove factual data. To mitigate bias and fairness issues, we recommend adding additional checks, such as those implemented in \citet{stowe-2024-identifying}, and involving human reviewers before rolling out any machine-generated content to learners.

We hope to show that a smaller model (i.e., a model that consumes less resources) can achieve the same performance as that of a larger model when training data is available. While smaller models are more accessible, they remain difficult to access in resource-limited environments where GPU compute is rare or expensive.
\section{Limitation}
The experiments were conducted for Llama-3 8B, and it is uncertain whether the findings will generalize to other models. The human evaluation of model performance was completed by the authors, who were also designed and conducted the experiment. As there may be unconscious biases on part of the authors, the dataset and annotations will be released upon acceptance.

\section*{Acknowledgements}

This document has been adapted
by Steven Bethard, Ryan Cotterell and Rui Yan
from the instructions for earlier ACL and NAACL proceedings, including those for 
ACL 2019 by Douwe Kiela and Ivan Vuli\'{c},
NAACL 2019 by Stephanie Lukin and Alla Roskovskaya, 
ACL 2018 by Shay Cohen, Kevin Gimpel, and Wei Lu, 
NAACL 2018 by Margaret Mitchell and Stephanie Lukin,
Bib\TeX{} suggestions for (NA)ACL 2017/2018 from Jason Eisner,
ACL 2017 by Dan Gildea and Min-Yen Kan, 
NAACL 2017 by Margaret Mitchell, 
ACL 2012 by Maggie Li and Michael White, 
ACL 2010 by Jing-Shin Chang and Philipp Koehn, 
ACL 2008 by Johanna D. Moore, Simone Teufel, James Allan, and Sadaoki Furui, 
ACL 2005 by Hwee Tou Ng and Kemal Oflazer, 
ACL 2002 by Eugene Charniak and Dekang Lin, 
and earlier ACL and EACL formats written by several people, including
John Chen, Henry S. Thompson and Donald Walker.
Additional elements were taken from the formatting instructions of the \emph{International Joint Conference on Artificial Intelligence} and the \emph{Conference on Computer Vision and Pattern Recognition}.

\bibliography{custom}

\clearpage
\onecolumn
\appendix
\section{Appendix}

\label{sec:appendix}

\subsection{Prompt templates}\label{sec:prompt-templates}
\subsubsection{Prompt Template to generate ELPA Instructions}\label{sec:prompt-template-1}

\begin{tcolorbox}[colback=gray!10,colframe=gray!30,coltitle=black,title=Generation prompt template]
\small 
You are asked to come up with a set of 15 task instructions in English. These instructions should be useful for language learners of English.
These task instructions will be given to a GPT model and we will evaluate the GPT model for completing the instructions. Separate each instruction using "\#\#\#". \\
Here are the requirements: \\
1. The type of instructions should be similar and related to the instructions in the prompt. \\
2. These instructions should be related English language learning, such as grammars, semantics, pragmatics, etc. \\
3. Please don't write instructions to write a code or program or answer a mathematical question. \\
4. Please avoid generating factual instructions that ask specific questions on history, geography, politics, or science. \\
5. The instructions should not contain racist, sexist, toxic, or otherwise potentially offensive language. \\
6. Not all instructions require input. For example, when an instruction asks "did you have lunch yet", it is not necessary to provide a specific context. In this case, we simply put "<noinput>" in the input field. \\
7. The output should be an appropriate response to the instruction and the input. \\
List of 15 tasks:
\\ \\
\textcolor{gray}{// Here we insert a 3 seed instructions and 1 model-generated example.\\
1. Instruction: \dots \\
1. Input: \dots \\
1. Output: \dots \\
1. Explanation: \dots
\\ 
\#\#\# \\
\dots
\\
\#\#\# \\
5. Instruction: \dots \\
5. Input: \dots \\
5. Output: \dots \\
5. Explanation: \dots 
}
\\ \#\#\# \\
6. Instruction: \\
\end{tcolorbox}

\newpage 
\subsubsection{ELPA Instruction Data Filtration}\label{sec:prompt-template-2}

\begin{tcolorbox}
[colback=gray!10,colframe=gray!30,coltitle=black,title=Filtration prompt template]
\small
Given the following tuples of <instruction,input,output> your task is to evaluate the quality of the
tuple(s) and accept or reject them based on the following requirements. Separate each example using ``\#\#\#''.
\\
Here are the requirements:
\\
1. The <instruction,input,output> tuples are used for language learning in English. \\
2. The output should not contain any verifiable factual information related to science, geography, history, business etc. \\
3. The output should not contain any code, program, or mathematical formula. \\
4. The output should be free of racist, sexist, toxic or otherwise potentially offensive language and imagery. \\
5. The instruction should not contain inflammatory, highly controversial or upsetting topics. \\
6. The output should not contain language or symbols that reinforce stereotypes. \\
7. Return the full tuple <instruction,input,output> with Evaluation and Reason. \\

See the following examples. \\
1. Instruction: Reply the statement with a proper answer. \\
Input: we are all going to the movie at 7pm. \\
Output: Great, my work finishes at 6pm so perhaps I can join too. \\
Evaluation: Accept. \\
Reason: The instruction, input, and the output does not contain any factual information, code, or inflammatory comment. \\
\#\#\# \\
2. Instruction: Write an indirect answer to the question. \\
Input: who is your favorite soccer player? \\
Output: I loathe soccer because it is not a manly sport. \\
Evaluation: Reject. \\
Reason: The output seems toxic ("not a manly sport") and gender biased. \\
\#\#\# \\
3. Instruction: This is an email written by a customer to a customer support team. Please give me a question that asks about the main idea. \\
Input: Email: Hello, Thanks for sending my order \#3397—it arrived this morning. Unfortunately, the paint was not the one I had asked for. I had selected color SP 944 but received SP 945 (Ocean Waves). They appear right next to each other on your Web site, so the two may have been confused at your end. Could you send me the correct paint, along with additional samples that are close in color to SP 722? Thank you, Arun Phan \\
Output: What problem does Mr. Phan mention in his e-mail? \\
Evaluation: Accept. \\
Reason: The instruction, input, and the output does not contain any factual information, code, or inflammatory comment. \\
\#\#\# \\
4. Instruction: Write a 6-turn exchange between 3 people (Person-1, Person-2, and Person-3). They all work at the same company, and discuss thoughts on which division will end up occupying the space. \\
Input: no-input. \\
Output: \\
"Person-1: Have you two taken a look at the progress they’ve made upstairs on the office expansion? It looks great! \\
Person-2: I know! I can’t believe it! And the offices up there have amazing views of the city. \\
Person-3: I wonder which division will move up there when it’s finished. \\
Person-2: I heard it’s the research department. \\
Person-1: Ah, because the CEO is biased towards the department. In fact the CEO hired her husband to lead a new project inside research. \\
Person-1: I think you’re right, there!" \\
Evaluation: Reject. \\
Reason: The output seems toxic. The conversation is not suitable for a workplace environment. \\
\#\#\# \\
5. Instruction: What is the capital of India? \\
Input: no-input. \\
Output: New Delhi. \\
Evaluation: Reject. \\
Reason: The output is factual.

\end{tcolorbox}

\newpage
\subsection{Model performance results}\label{subsection:win-tie-rate}
\newcolumntype{P}[1]{>{\centering\arraybackslash}p{#1}}

\begin{table*}[!htbp]
    \centering
    \small
    \begin{tabular}{c|P{1.cm}|P{1.cm}|P{1.cm}|P{1.cm}|P{1.cm}|P{1.cm}|P{1.cm}|P{1.cm}}
         &  \rotatebox{60}{Dolly} & \rotatebox{60}{Mistral} & \rotatebox{60}{GPT-3.5} & \rotatebox{60}{Llama} & \rotatebox{60}{SFT-17K} & \rotatebox{60}{SFT-50K} & \rotatebox{60}{SFT-70K} \\
        \hline 
        Dolly &  \cellcolor{gray!25} & \textcolor{darkgray}{8.50} \textcolor{olive}{4.50} \textcolor{magenta}{1.50}  & \textcolor{darkgray}{4.00} \textcolor{olive}{2.00} \textcolor{magenta}{2.00}  & \textcolor{darkgray}{15.50} \textcolor{olive}{14.50} \textcolor{magenta}{3.50}  & \textcolor{darkgray}{7.50} \textcolor{olive}{4.00} \textcolor{magenta}{1.00}  & \textcolor{darkgray}{6.00} \textcolor{olive}{3.00} \textcolor{magenta}{0.00}  & \textcolor{darkgray}{4.50} \textcolor{olive}{2.00} \textcolor{magenta}{0.00}  \\
        \hline 
        Mistral & \textcolor{darkgray}{71.00} \textcolor{olive}{59.50} \textcolor{magenta}{69.50}  & \cellcolor{gray!25} & \textcolor{darkgray}{20.00} \textcolor{olive}{9.00} \textcolor{magenta}{35.00}  & \textcolor{darkgray}{54.00} \textcolor{olive}{42.00} \textcolor{magenta}{44.50}  & \textcolor{darkgray}{25.50} \textcolor{olive}{14.50} \textcolor{magenta}{16.50}  & \textcolor{darkgray}{23.00} \textcolor{olive}{11.00} \textcolor{magenta}{12.00}  & \textcolor{darkgray}{24.00} \textcolor{olive}{10.50} \textcolor{magenta}{13.00}  \\
        \hline 
        GPT-3.5 & \textcolor{darkgray}{72.50} \textcolor{olive}{60.00} \textcolor{magenta}{51.00}  & \textcolor{darkgray}{34.50} \textcolor{olive}{12.00} \textcolor{magenta}{29.50}  & \cellcolor{gray!25} & \textcolor{darkgray}{56.50} \textcolor{olive}{43.00} \textcolor{magenta}{37.00}  & \textcolor{darkgray}{25.50} \textcolor{olive}{13.50} \textcolor{magenta}{14.50}  & \textcolor{darkgray}{27.50} \textcolor{olive}{11.50} \textcolor{magenta}{11.50}  & \textcolor{darkgray}{22.50} \textcolor{olive}{11.00} \textcolor{magenta}{12.00}  \\
        \hline
        Llama & \textcolor{darkgray}{38.00} \textcolor{olive}{32.50} \textcolor{magenta}{46.50}  & \textcolor{darkgray}{15.50} \textcolor{olive}{5.00} \textcolor{magenta}{21.00}  & \textcolor{darkgray}{8.00} \textcolor{olive}{3.00} \textcolor{magenta}{23.00}  & \cellcolor{gray!25} & \textcolor{darkgray}{12.00} \textcolor{olive}{6.50} \textcolor{magenta}{8.50}  & \textcolor{darkgray}{10.50} \textcolor{olive}{4.50} \textcolor{magenta}{5.00}  & \textcolor{darkgray}{7.00} \textcolor{olive}{4.50} \textcolor{magenta}{5.00}  \\
        \hline
        SFT-17K &\textcolor{darkgray}{68.50} \textcolor{olive}{55.00} \textcolor{magenta}{76.50}  & \textcolor{darkgray}{33.00} \textcolor{olive}{10.50} \textcolor{magenta}{44.50}  & \textcolor{darkgray}{18.50} \textcolor{olive}{6.50} \textcolor{magenta}{39.50}  & \textcolor{darkgray}{55.00} \textcolor{olive}{39.50} \textcolor{magenta}{54.50}  & \cellcolor{gray!25} & \textcolor{darkgray}{24.00} \textcolor{olive}{10.50} \textcolor{magenta}{12.50}  & \textcolor{darkgray}{18.00} \textcolor{olive}{8.50} \textcolor{magenta}{10.50}  \\
        \hline
        SFT-50K & \textcolor{darkgray}{72.00} \textcolor{olive}{58.50} \textcolor{magenta}{81.50}  & \textcolor{darkgray}{32.50} \textcolor{olive}{11.50} \textcolor{magenta}{51.50}  & \textcolor{darkgray}{22.50} \textcolor{olive}{9.00} \textcolor{magenta}{46.50}  & \textcolor{darkgray}{54.50} \textcolor{olive}{42.00} \textcolor{magenta}{58.50}  & \textcolor{darkgray}{26.00} \textcolor{olive}{15.00} \textcolor{magenta}{21.00}  & \cellcolor{gray!25} & \textcolor{darkgray}{15.50} \textcolor{olive}{8.00} \textcolor{magenta}{11.00}  \\
        \hline
        SFT-70K & \textcolor{darkgray}{70.50} \textcolor{olive}{59.00} \textcolor{magenta}{85.50}  & \textcolor{darkgray}{39.00} \textcolor{olive}{12.50} \textcolor{magenta}{56.00}  & \textcolor{darkgray}{23.50} \textcolor{olive}{10.00} \textcolor{magenta}{50.00}  & \textcolor{darkgray}{56.50} \textcolor{olive}{43.50} \textcolor{magenta}{60.50}  & \textcolor{darkgray}{26.00} \textcolor{olive}{14.50} \textcolor{magenta}{25.00}  & \textcolor{darkgray}{22.50} \textcolor{olive}{9.50} \textcolor{magenta}{16.50}  & \cellcolor{gray!25} \\
    \hline 
    \end{tabular}
    \caption{The table represents the \textbf{win-rate} of each model on the y-axis when compared to the model on the x-axis. Validity in \textcolor{darkgray}{dark gray}, Output correctness in \textcolor{olive}{olive}, and Quality of explanation in \textcolor{magenta}{magenta}.
    }
    \label{tab:my_label}
\end{table*}

\begin{table*}[!htbp]
    \centering
    \small
    \begin{tabular}{c|P{1.cm}|P{1.cm}|P{1.cm}|P{1.cm}|P{1.cm}|P{1.cm}|P{1.cm}|P{1.cm}}
         &  \rotatebox{60}{Dolly} & \rotatebox{60}{Mistral} & \rotatebox{60}{GPT-3.5} & \rotatebox{60}{Llama} & \rotatebox{60}{SFT-17K} & \rotatebox{60}{SFT-50K} & \rotatebox{60}{SFT-70K} \\
        \hline 
        Dolly &  \cellcolor{gray!25} & \textcolor{darkgray}{20.50} \textcolor{olive}{36.00} \textcolor{magenta}{29.00}  & \textcolor{darkgray}{23.50} \textcolor{olive}{38.00} \textcolor{magenta}{47.00}  & \textcolor{darkgray}{46.50} \textcolor{olive}{53.00} \textcolor{magenta}{50.00}  & \textcolor{darkgray}{24.00} \textcolor{olive}{41.00} \textcolor{magenta}{22.50}  & \textcolor{darkgray}{22.00} \textcolor{olive}{38.50} \textcolor{magenta}{18.50}  & \textcolor{darkgray}{25.00} \textcolor{olive}{39.00} \textcolor{magenta}{14.50}   \\
        \hline 
        Mistral & \cellcolor{gray!25} & \cellcolor{gray!25} & \textcolor{darkgray}{45.50} \textcolor{olive}{79.00} \textcolor{magenta}{35.50}  & \textcolor{darkgray}{30.50} \textcolor{olive}{53.00} \textcolor{magenta}{34.50}  & \textcolor{darkgray}{41.50} \textcolor{olive}{75.00} \textcolor{magenta}{39.00}  & \textcolor{darkgray}{44.50} \textcolor{olive}{77.50} \textcolor{magenta}{36.50}  & \textcolor{darkgray}{37.00} \textcolor{olive}{77.00} \textcolor{magenta}{31.00}   \\
        \hline 
        GPT-3.5 & \cellcolor{gray!25} & \cellcolor{gray!25} & \cellcolor{gray!25} & \textcolor{darkgray}{35.50} \textcolor{olive}{54.00} \textcolor{magenta}{40.00}  & \textcolor{darkgray}{56.00} \textcolor{olive}{80.00} \textcolor{magenta}{46.00}  & \textcolor{darkgray}{50.00} \textcolor{olive}{79.50} \textcolor{magenta}{42.00}  & \textcolor{darkgray}{54.00} \textcolor{olive}{79.00} \textcolor{magenta}{38.00}   \\
        \hline
        Llama & \cellcolor{gray!25} & \cellcolor{gray!25} & \cellcolor{gray!25} & \cellcolor{gray!25} & \textcolor{darkgray}{33.00} \textcolor{olive}{54.00} \textcolor{magenta}{37.00}  & \textcolor{darkgray}{35.00} \textcolor{olive}{53.50} \textcolor{magenta}{36.50}  & \textcolor{darkgray}{36.50} \textcolor{olive}{52.00} \textcolor{magenta}{34.50}  \\
        \hline
        SFT-17K & \cellcolor{gray!25} & \cellcolor{gray!25} & \cellcolor{gray!25} & \cellcolor{gray!25} & \cellcolor{gray!25} & \textcolor{darkgray}{50.00} \textcolor{olive}{74.50} \textcolor{magenta}{66.50}  & \textcolor{darkgray}{56.00} \textcolor{olive}{77.00} \textcolor{magenta}{64.50}   \\
        \hline
        SFT-50K &\cellcolor{gray!25} & \cellcolor{gray!25} & \cellcolor{gray!25} & \cellcolor{gray!25} & \cellcolor{gray!25} & \cellcolor{gray!25} & \textcolor{darkgray}{62.00} \textcolor{olive}{82.50} \textcolor{magenta}{72.50}   \\
        \hline
    \end{tabular}
    \caption{The table represents the \textbf{tie-rate} of each model on the y-axis when compared to the model on the x-axis. Validity in \textcolor{darkgray}{dark gray}, Output correctness in \textcolor{olive}{olive}, and Quality of explanation in \textcolor{magenta}{magenta}.}
    \label{tab:my_label}
\end{table*}

\newpage
\subsection{Full evaluation examples}

\subsubsection{Example with a formatting error in the output}\label{format-example}
\begin{itemize}
    \item \textbf{Instruction}: Generate a question to assess understanding of the main point in the following sentence.
    \item \textbf{Input}: In spite of facing hardships, the team managed to win the championship.
\end{itemize}
\begin{table*}[h!]
    \centering
    \small
    \begin{tabular}{|P{4.5cm}|l|P{4.5cm}|P{1.5cm}|P{1.5cm}|P{1.5cm}|}
    \hline
        \textbf{Output} & \textbf{Model} & \textbf{Explanation} & \textbf{Validity} & \textbf{Output \newline correctness} & \textbf{Quality of \newline explanation} \\ \hline
       The team generated a question to assess their understanding of the main point in the above sentence - In spite of facing hardships, the team managed to win the championship.&dolly& &invalid&wrong&no\\ \hline
       What did the team manage to do in spite of facing hardships?&gpt& &valid and ready for a test&right&no\\ \hline
       What challenge(s) did the team encounter while winning the championship?&mistral&The question tests understanding of the main point by asking about the challenges the team faced while achieving their victory. The sentence implies that the team's success was not easy but required overcoming obstacles, so asking about these challenges directly assesses comprehension of the sentence's meaning.&valid and ready for a test&right&yes\\ \hline
       1. Why did the team win the championship in spite of facing hardships? 2. Why did the team win the championship despite facing hardships? 3. Why did the team win the championship despite the hardships they faced? 4. Why did the team win the championship despite the hardships they faced? 5. Why did the team win the championship despite the hardships they faced? 6. Why did the team win the championship despite the hardships they faced? 7. Why did the team win the championship despite the hardships they faced? 8. Why did the team win the championship despite the hardships they faced? 9. Why did the team win the championship despite the hardships they faced? 10. Why did the team win the championship despite the hardships they faced?&llama& The question is appropriate because it assesses the main point of the sentence, which is that the team won the championship despite facing hardships. The question also encourages the student to think critically about the reasons why the team was able to win the championship in spite of the hardships they faced.&invalid&wrong&no\\ \hline
       What was the outcome of the team despite the hardships?  &sft-17k& The question is designed to assess whether the respondent understands the main point of the sentence, which is the team's success in spite of hardships.&valid&right&weak yes\\ \hline
       What was the outcome of the team's situation despite facing hardships?  &sft-50k& The question directly asks for the main point or outcome of the situation described in the input sentence, which is that the team won the championship despite facing hardships.&valid&right&weak yes\\ \hline
       What is the main point that the sentence is trying to convey about the team's victory?  &sft-70k& The question is designed to assess the reader's understanding of the main point of the sentence, which is the team's victory despite hardships.&valid&right&weak yes\\ \hline
    \end{tabular}
\end{table*}

\newpage
\subsubsection{Example where output is in the proximity but does not follow the instruction exactly}\label{proximity-example}
\begin{itemize}
    \item \textbf{Instruction}: Paraphrase the following sentence.
    \item \textbf{Input}: Nobody knew how much time she spent training for the Olympic Games.
\end{itemize}
\begin{table*}[h!]
    \centering
    \small
    \begin{tabular}{|P{4.5cm}|l|P{4.5cm}|P{1.5cm}|P{1.5cm}|P{1.5cm}|}
    \hline
        \textbf{Output} & \textbf{Model} & \textbf{Explanation} & \textbf{Validity} & \textbf{Output \newline correctness} & \textbf{Quality of \newline explanation} \\ \hline
        Sara Performance, the Canadian freestyle skier who competed in the 2006 and 2010 Olympic Games, spent three years training for the games. No one knew how much time she spent training because she kept it a secret to avoid pressure from expectations.&dolly& & invalid & wrong & no \\ \hline
        No one was aware of the amount of time she dedicated to training for the Olympic Games.&gpt & The output is appropriate given the instruction and input because it rephrases the original sentence while retaining the meaning. It replaces "nobody knew" with "no one was aware of" and "how much time" with "the amount of time". Additionally, it changes "spent training" to "dedicated to training" to convey the same idea. & valid & right & yes \\ \hline
        It was unknown how many hours she dedicated to preparing for the Olympic Games. \newline This paraphrase conveys the same meaning as the original sentence but uses different words and phrase structures. The original sentence says "Nobody knew," which can be paraphrased as "It was unknown." The instruction also asks to change "she spent" to "she dedicated," which creates a more formal and specific phrasing. The phrase "for the Olympic Games" remains the same in both sentences. By using "hours" instead of "time," the paraphrase provides a clearer understanding of the quantity of training time being referred to. &mistral& &valid & right & weak yes \\ \hline
        She trained a lot for the Olympic Games.&llama & The output is a paraphrase of the input sentence. It is appropriate given the instruction to paraphrase the sentence. & invalid & wrong & no  \\ \hline
        The extent of her training for the Olympics remained a mystery to everyone.&sft-17k & The paraphrased sentence effectively conveys the same meaning as the original sentence but uses different words and syntax to do so. & valid and ready for a test & right & yes  \\ \hline
        Nobody was aware of the amount of time she spent training for the Olympic Games.&sft-50k & The reply paraphrases the original sentence by replacing 'knew' with 'was aware of' and 'how much time' with 'the amount of time', while keeping the same meaning. & valid & right & weak yes \\ \hline & 
        sft-70k&&invalid & wrong & no \\ \hline
    \end{tabular}
\end{table*}

\newpage 
\subsubsection{Example where the task is easy for most models}\label{easy-example}
\begin{itemize}
    \item \textbf{Instruction}: Translate the given English sentence into simple words that a beginner might understand.
    \item \textbf{Input}: The CEO convened an emergency meeting to discuss the catastrophic consequences of the failed project.
\end{itemize}

\begin{table*}[h!]
    \centering
    \small
    \begin{tabular}{|P{4.5cm}|l|P{4.5cm}|P{1.5cm}|P{1.5cm}|P{1.5cm}|}
    \hline
        \textbf{Output} & \textbf{Model} & \textbf{Explanation} & \textbf{Validity} & \textbf{Output \newline correctness} & \textbf{Quality of \newline explanation} \\ \hline
        discuss catastrophic consequences of failed project & dolly & & invalid & wrong & no \\ \hline

        1. Boss gathered people, 2. Quick meeting, 3. Discuss bad project problem. & mistral & In this sentence, "The CEO convened an emergency meeting to discuss the catastrophic consequences of the failed project", the word 'CEO' refers to the leader or boss of a company, 'convened' means called or gathered, 'emergency' indicates that it's important and urgent, 'discuss' means talking about a problem, and 'catastrophic consequences' refer to the serious negative outcomes. To simplify it for a beginner, we can break down the sentence into three parts: 1. Boss gathered people, 2. Quick meeting, 3. Discuss bad project problem. This way, a beginner can understand the sentence without getting overwhelmed by long words and complex sentences. & valid & right & weak yes \\ \hline 

        The boss gathered everyone for a sudden meeting to talk about the really bad results of the project that did not go well. & gpt & & valid and ready for a test & right & no \\ \hline 

        1. The CEO called a meeting to talk about the bad results of the project.  & llama &  The CEO convened an emergency meeting to discuss the catastrophic consequences of the failed project. This means that the CEO called a meeting to talk about the bad results of the project. The output is appropriate because it uses simple words that a beginner might understand and it accurately reflects the meaning of the input sentence. & valid and ready for a test & wrong & weak yes \\ \hline 

        The boss called a meeting quickly to talk about the serious problems with a project that didn't work.  & sft-17k &  The reply translates the high-level English in the input to simpler, beginner-level words without changing the overall meaning of the sentence. & valid and ready for a test & right & yes \\ \hline 

        The boss called a very important meeting to talk about why a project didn't work well.   & sft-50k &  The output provides a simpler version of the input, using easier words and shorter phrases while retaining the original meaning. It is appropriate for a beginner to understand. & valid and ready for a test & right & yes \\ \hline 

        The top boss called a meeting to talk about the bad results of a failed project.  & sft-70k &  The reply simplifies the language of the original sentence by using simpler words and phrases like 'top boss' instead of 'CEO', 'called a meeting' instead of 'convened an emergency meeting', and 'talk about the bad results' instead of 'discuss the catastrophic consequences'. & valid and ready for a test & right & yes \\ \hline
    \end{tabular}
\end{table*}

\newpage
\subsubsection{Example where the task is difficult for most models}\label{hard-example}
\begin{itemize}
    \item \textbf{Instruction}: Write a sentence using an idiomatic expression.
    \item \textbf{Input}: 
\end{itemize}

\begin{table*}[h!]
    \centering
    \small
    \begin{tabular}{|P{4.5cm}|l|P{4.5cm}|P{1.5cm}|P{1.5cm}|P{1.5cm}|}
    \hline
        \textbf{Output} & \textbf{Model} & \textbf{Explanation} & \textbf{Validity} & \textbf{Output \newline correctness} & \textbf{Quality of \newline explanation} \\ \hline

        noinput.&dolly& &invalid&wrong&no\\ \hline

        The project is a 'pie in the sky' idea that may never see the light of day.&mistral& &valid and ready for a test&right&yes\\ \hline

       No pain, no gain.&gpt& - &valid&right&wrong\\ \hline

       The first input is a sentence, the second a sequence of words that together form an idiomatic expression.\#\# Prompt
        Prompt: Make an idiomatic expression out of the first input sentence.
        \#\# Difficulty \dots
  &llama& &invalid&wrong&no\\ \hline

         After hours of studying, he felt like he was burning the midnight oil.&sft-17k& The phrase 'burning the midnight oil' is an idiomatic expression meaning to work late into the night. In this context, it is used to depict the act of studying late into the night.&valid and ready for a test&right&yes\\ \hline

         &sft-50k&  An idiomatic expression or idiom is a phrase where the words together have a meaning that is different from the dictionary definitions of the individual words. An example of an idiom is 'it's raining cats and dogs', which means it's raining heavily.&invalid&wrong&no\\ \hline

         It's raining cats and dogs outside.&sft-70k& The sentence uses the idiomatic expression 'raining cats and dogs', which means it's raining heavily.&valid and ready for a test&right&yes\\ \hline
    \end{tabular}
\end{table*}

\end{document}